%% file: 0-main.tex
\documentclass[sigconf]{acmart}

\usepackage{url}
\usepackage{threeparttable}
\usepackage{amsmath,amssymb,amsfonts,multirow,algorithm,algorithmic,float}
\usepackage{algorithmic}
\usepackage{graphicx}
\usepackage{subfigure}
\usepackage{textcomp}
\usepackage{xcolor}
\usepackage{booktabs}
\usepackage{setspace} 

\def\BibTeX{{\rm B\kern-.05em{\sc i\kern-.025em b}\kern-.08emT\kern-.1667em\lower.7ex\hbox{E}\kern-.125emX}}

\def\methodname{\texttt{MetaPred}}

\copyrightyear{2019}
\acmYear{2019}
\setcopyright{acmcopyright}
\acmConference[KDD '19] {The 25th ACM SIGKDD Conference on Knowledge Discovery and Data Mining}{August 4--8, 2019}{Anchorage, AK, USA}
\acmBooktitle{The 25th ACM SIGKDD Conference on Knowledge Discovery and Data Mining (KDD'19), June 22--24, 2019, Anchorage, AK, USA}
\acmPrice{15.00}
\acmDOI{10.1145/XXXXXX.XXXXXX}
\acmISBN{978-1-4503-6201-6/19/08} 

\begin{document}
\title[]{MetaPred: Meta-Learning for Clinical Risk Prediction with Limited Patient Electronic Health Records}

\author{Xi Sheryl Zhang$^1$, Fengyi Tang$^2$, Hiroko Dodge$^{3,4}$, Jiayu Zhou$^2$, Fei Wang$^1$}
\orcid{}
\affiliation{%
  \institution{$^1$Department of Healthcare Policy and Research. Weill Cornell Medicine. Cornell University.\\
  $^2$Department of Computer Science and Engineering. Michigan State University.\\
  $^3$Michigan Alzheimer's Disease Center, Department of Neurology, University of Michigan.\\
  $^4$Layton Aging and Alzheimer's Disease Center, Department of Neurology, Oregon Health \& Science University.}
  \streetaddress{}
  \city{}
  \state{}
  \postcode{}
}
\email{sheryl.zhangxi@gmail.com;{tangfeng,jiayuz}@msu.edu;hdodge@med.umich.edu;few2001@med.cornell.edu}

\begin{abstract}
In recent years, increasingly augmentation of health data, such as patient Electronic Health Records (EHR), are becoming readily available. This provides an unprecedented opportunity for knowledge discovery and data mining algorithms to dig insights from them, which can, later on, be helpful to the improvement of the quality of care delivery. Predictive modeling of clinical risk, including in-hospital mortality, hospital readmission, chronic disease onset, condition exacerbation, etc., from patient EHR, is one of the health data analytic problems that attract most of the interests. The reason is not only because the problem is important in clinical settings, but also there are challenges working with EHR such as sparsity, irregularity, temporality, etc. Different from applications in other domains such as computer vision and natural language processing, the labeled data samples in medicine (patients) are relatively limited, which creates lots of troubles for effective predictive model learning, especially for complicated models such as deep learning. In this paper, we propose~\texttt{MetaPred}, a meta-learning for clinical risk prediction from longitudinal patient EHRs. In particular, in order to predict the target risk where there are limited data samples, we train a meta-learner from a set of related risk prediction tasks which learns how a good predictor is learned. The meta-learned can then be directly used in target risk prediction, and the limited available samples can be used for further fine-tuning the model performance. The effectiveness of \texttt{MetaPred} is tested on a real patient EHR repository from Oregon Health \& Science University. We are able to demonstrate that with Convolutional Neural Network (CNN) and Recurrent Neural Network (RNN) as base predictors, \texttt{MetaPred} can achieve much better performance for predicting target risk with low resources comparing with the predictor trained on the limited samples available for this risk. 
\end{abstract}

%
%


\keywords{meta-learning, clinical risk prediction, electronic health records}

\maketitle

\input{1-introduction}
\input{2-problemsetup}
\input{3-method}
\input{4-experiment}

\input{5-relatedwork}
\input{6-conclusion}

\input{7-acknowledgment}

\begin{spacing}{1} 
\bibliographystyle{ACM-Reference-Format}
\bibliography{myrefs}\tiny
\end{spacing}

\end{document}

%% file: 1-introduction.tex
\section{Introduction}
The recent years have witnessed a surge of interests in healthcare analytics with longitudinal patient Electronic Health Records (EHR) \cite{jensen2012mining}. Predictive modeling of clinical risk, such as mortality \cite{sun2018identify,tang2018predictive}, hospital readmission \cite{caruana2015intelligible,shadmi2015predicting}, onset of chronic disease \cite{choi2016using}, condition exacerbation \cite{kerkhof2015predicting}, etc., has been one of the most popular research topics. This is mainly because 1) accurate clinical risk prediction models can help the clinical decision makers to identify the potential risk at its early stage, therefore appropriate actions can be taken in time to provide the patient with better care; 2) there are many challenges on analyzing patient EHR, such as sequentiality, sparsity, noisiness, irregularity, etc. \cite{wang2012towards}. Many computational algorithms have been developed to overcome these challenges, including both conventional approaches \cite{caruana2015intelligible} and deep learning models \cite{choi2016using}.

One important characteristic that makes those healthcare problems different from the applications in other domains, such as computer vision \cite{lecun2015deep}, speech analysis \cite{deng2013recent} and natural language processing \cite{young2018recent}, is that the number of the available sample data set is typically limited, and typically it is very expensive and sometimes even impossible for obtaining new samples. For example, for the case of individualized patient risk prediction, where the goal is to predict a certain clinical risk for each patient, each data sample corresponds to a patient. There are in total just 7.5 billion people all over the world, and the number will be far less if we focus on a specific disease condition. These patients are also distributed in different continents, different states, different cities, and different hospitals. The reality is that we only have a small number of patients available in a specific EHR corpus for training a risk prediction model. Moreover, the clinical risks we focus on are extraordinarily complicated. For the majority of the deadly diseases, we are still not clear about their underlying biological mechanisms and thus the potential treatment strategies. This means that, in order to learn accurate clinical risk prediction models, we need to make sufficient use the limited patient samples, and effectively leverage available knowledge about the clinical risk as well as predictive models.

Recently, transfer learning \cite{pan2010survey} has been demonstrated as an effective mechanism to achieve good performance in learning with limited samples in medical problems. For example, in computer vision, Inception-V3 \cite{szegedy2016rethinking} is a powerful model for image analysis. Google has released the model parameters trained on the huge ImageNet data set \cite{deng2009imagenet}. Esteva {\em et al.} \cite{esteva2017dermatologist} adopted such a model as the starting point, and leveraged a locally collected 130K skin images to fine-tune the model to discriminate benign vs. malignant skin lesions. They achieved satisfactory classification performance that is comparable to the performance of well-trained dermatologists. Similar strategies have also achieved good performance in other medical problems with different types of medical images \cite{gulshan2016development,kermany2018identifying}. In addition to computer vision, powerful natural language processing models such as transformer \cite{vaswani2017attention} and BERT \cite{devlin2018bert} with parameters trained on general natural language data, have also been fine-tuned to analyze unstructured medical data \cite{lee2019biobert}. Because these models are pre-trained on general data, they can only encode some general knowledge, which is not specific to medical problems. Moreover, such models are only available with certain complicated architectures with a huge amount of general training data. It is difficult to judge how and why such a mechanism will be effective in which clinical scenarios.

In this paper, we propose~\texttt{MetaPred}, a meta-learning framework for low-resource predictive modeling with patient EHRs. Meta-learning \cite{thrun1998learning, ritter2018been} is a recent trend in machine learning aiming at learning to learn. By low-resource, we mean that only limited EHRs can be used for the target clinical risk, which is insufficient to train a good predictor by seen samples of the task themselves. For this scenario, we develop a model agnostic gradient descent framework to train a meta-learner on a set of prediction tasks where the target clinical risks are highly relevant. For these tasks, we choose one of them as the simulated target and the rest as sources. The parameters of the predictive model will be updated through a step-by-step sequential optimization process. In each step, an {\em episode} of data will be sampled from the sources and the simulated target to support the updating on model parameters. To compensate for the optimization-level fast adaptation, an objective-level adaptation is also proposed. We validate the effectiveness of~\texttt{MetaPred} on a large-scale real-world patient EHR corpus with a set of cognition related disorders as the clinical risks to be predicted, and Convolutional Neural Networks (CNN) as well as Long-Short Term Memory (LSTM) are applied as the predictors because of their popularity in EHR-based analysis. Additionally, we demonstrate that if we use EHRs in target domains to fine-tune the learned model, the prediction performance can be further improved.

The rest of the paper is organized as follows: the problem setup is presented in Section 2; the proposed framework \texttt{MetaPred} is introduced in Section 3; experimental results are shown in Section 4 and related works are summarized in Section 5; finally, conclusion reaches at Section 6.

%% file: 2-problemsetup.tex
\section{Problem Setup}

\begin{figure}[htb]
    \centering
    \includegraphics[width=3.5in]{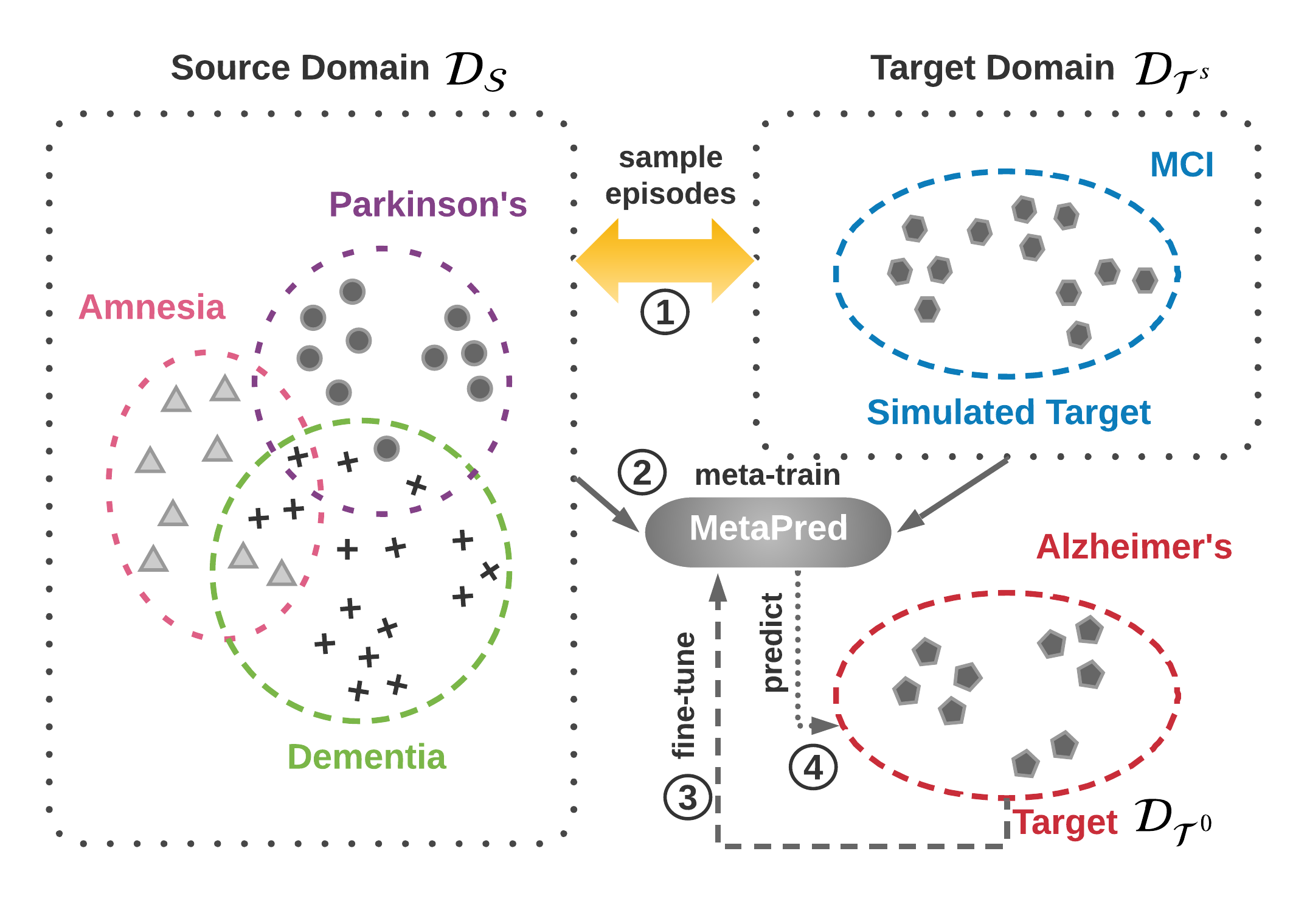}
    \caption{Illustration of the proposed learning procedure. In this example, our goal is to predict the risks of Alzheimer's disease with few labeled patients, which give rise to a low-resource classification. The idea is to take advantage of labeled patients from other relevant high-resource domains and design the learning to transfer workflow with sources and a simulated target via meta-learning.}
    \label{fig:overall}
\vspace{-0.5cm}
\end{figure}

In order to introduce our framework, we provide a graphical illustration in Figure \ref{fig:overall}. 
Suppose the target task is the prediction of the onset risk of Alzheimer's Disease where we do not have enough training patient samples, and we want to transfer knowledge from other related disease domains with sufficient labels such as Mild Cognitive Impairment (MCI) or Dementia. 
However, traditional transfer learning would be also constrained by the small number of training samples, especially for those with complicated neural networks. Consequently, we take advantage of meta-learning by setting a simulated target domain for learning to transfer. Though applying meta-learning settings on the top of low-resource medical records for disease prediction seems intuitive, how to set up the problem is crucial. 

More formally, we consider multiple related disease conditions as the set of source domains ${\mathcal{S}^{1}, \cdots, \mathcal{S}^{K}}$ and a target domain ${\mathcal{T}^0}$. This leads to $K+1$ domains in total. In each domain, we can construct a training data set including the EHRs of both case (positive) and control (negative) patients. 
We use the data collection $\{(\mathbf{X}, \mathbf{y})\}_{i}, i=0,1,\cdots,K$ to denote the features and labels of the patients in these $K+1$ domains. Our goal is to learn a predictive model $f$ for the target domain ${\mathcal{T}^0}$. In the following we use $\Theta$ to denote the parameters of $f$. 
Because only a limited number of samples are available in $\mathcal{T}^0$, we hope to leverage the data from those source domains, i.e., $f=(\mathcal{D}_{\mathcal{S}}, \mathbf{X}; \Theta)$, where $\mathcal{D}_{S}$ denotes the collection of data samples in the source domains. From the perspective of domain adaptation \cite{ben2010theory}, the problem can be reduced to the design and optimization of model $f$ in an appropriate form of $\mathcal{D}_{\mathcal{S}}$.            

In this section we will mainly introduce how to utilize the source domain data $\mathcal{D}_{\mathcal{S}}$ in our~\methodname~framework. The details on the design of $f$ will be introduced in the next section. In general, supervised meta-learning provides models trained by data episodes $\{\mathcal{D}_{i}\}$ which is composed of multiple samples. Each $\mathcal{D}_{i}$ is usually split into two parts according to their labels. 
We further refer to the domain where the testing data are from the {\em simulated target domain} $\mathcal{D}_{\mathcal{T}^s}$, and it is still one of the source domains. Followed previous work \cite{finn2017model, ravi2016optimization}, we called the training procedure based on this split as \textit{meta-train}, and the testing procedure as \textit{meta-test}. 


In summary, the proposed~\methodname~framework illustrated in Figure \ref{fig:overall} consists of four steps: (1) constructing episodes by sampling from the source domains and the simulated target domain; (2) learn the parameters of predictors in an episode-by-episode manner; (3) fine-tuning the model parameters on the genuine target domain; (4) predicting the target clinical risk.  

\begin{figure*}[t]
    \centering
    \includegraphics[width=6in]{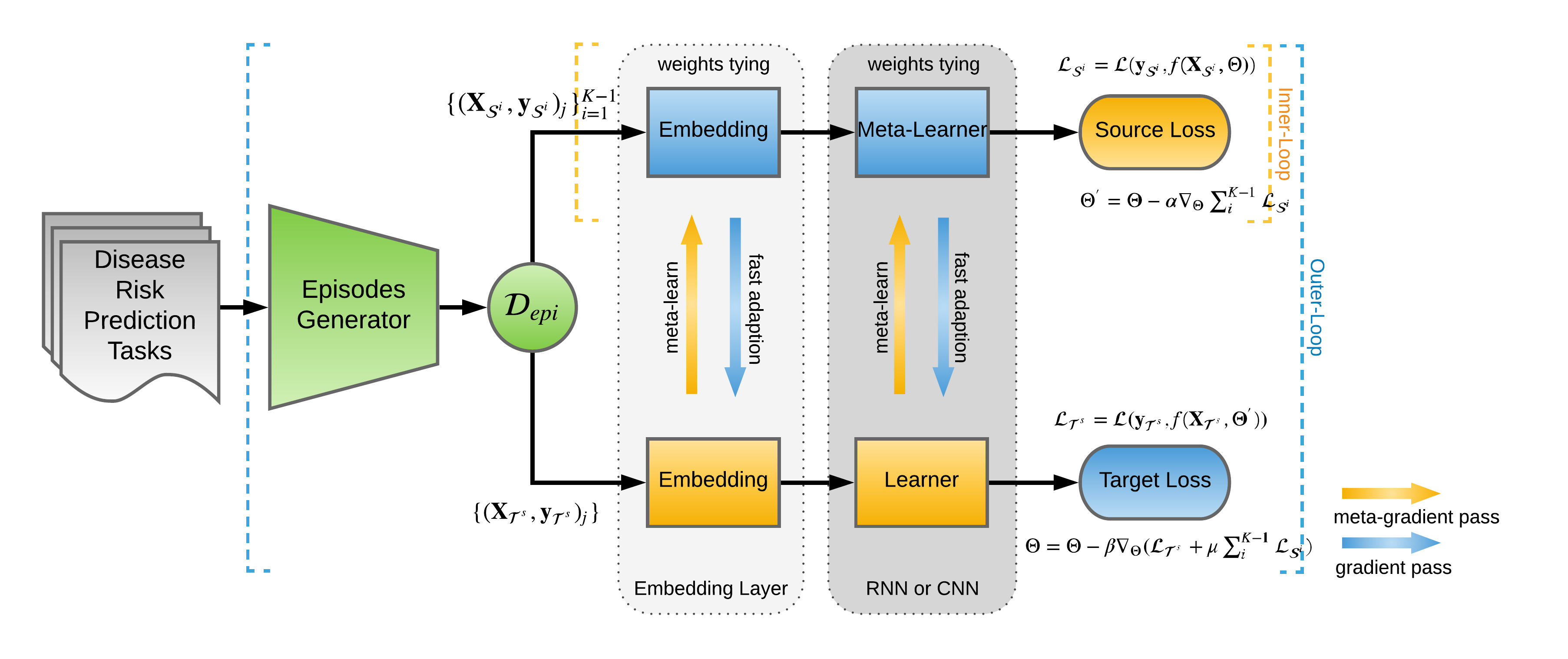}
    \caption{The overview of \methodname workflow. $\mathcal{D}_{epi}$ is an episode randomly sampled. $\{\mathcal{S}^i\}_{i=1}^{K-1}$ denotes source domains and $\mathcal{T}^s$ denotes the simulated target domain. The two gradient update loops of meta-training process are illustrated. The \textcolor{orange}{yellow} colored blocks and arrows are associated with Learner, while the \textcolor{blue}{blue} ones are associated with MetaLearner. (``Target loss'' is used here instead of ``Simulated Target loss'' for simplicity.)}
    \label{fig:Meta-SeqLearn}
\end{figure*}

%% file: 3-method.tex
\section{The \texttt{MetaPred} Framework}

The model-agnostic meta-learning strategy \cite{finn2017model} serves as the backbone of our~\methodname~framework. 
In particular, our goal is to learn a risk predictor on the target domain. In order to achieve that, we first perform model agnostic meta-learning on the source domains, where the model parameters are learned through
\begin{equation}
\begin{aligned}
\Theta^* = \mathbf{Learner}({\mathcal{T}^s}; \mathbf{MetaLearner}({\mathcal{S}^{1}, \cdots, \mathcal{S}^{K-1}}))
\end{aligned}
\label{eq:metalearn}
\end{equation}
where for each data episode, the model parameters are first adjusted through gradient descents on the objective loss measure on the training data from the source domains ($\mathbf{MetaLearner}$), and then they will be further adjusted on the simulated target domain $\mathcal{T}^s$ ($\mathbf{Learner}$). In the following, we will introduce the learning process in detail, where the risk prediction model is assumed to be either CNN or LTSM.
First we provide basic neural network prediction models as the options for $\mathbf{Learner}$. Then we introduce the entire parameter learning procedure of the proposed \methodname, including optimization-level adaptation and objective-level adaptation.

\subsection{Risk Prediction Models}

The EHR can be represented by sequences with multiple visits occurring at different time points for each patient. At each visit, the records can be depicted by a binary vector $\mathbf{x}^{t}\in\{0, 1\}^{\mathcal{|C|}}$, where $t$ denotes the time point. The values of $1$ indicate the corresponding medical event occurs at $t$, and 0 otherwise. $\mathcal{C}$ is the vocabulary of medical events, and $\mathcal{|C|}$ is its cardinality. Thus input of the predictive models can be denoted as a multivariate time series matrix $\mathbf{X}_{i} = \{\mathbf{x}_{i}^{t}\}_{t=1}^{T_i}$, where $i$ is the patient index and $T_i$ is the number of visits for patient $i$. The risk prediction model is trained to find a transformation mapping from input time series matrix $\mathbf{X}_i$ to the target disease label $\mathbf{y}_{i}\in \{0,1\}^2$. This makes the problem a sequence classification problem. 

\noindent{\bf {CNN-based Sequence Learning}}. There are three basic modules in our CNN based structure: embedding Layer, convolutional layer and multi-layer perceptron (MLP). Similar to natural language processing tasks \cite{kim2014convolutional}, 1-dimensional convolution operators are used to discover the data patterns along the temporal dimension $t$. Because the values of medical records at the visits are distributed in a discrete space, which is sparse and high-dimensional. It is necessary to place an embedding layer before CNN, to obtain a more compact continuous space for patient representation. The learnable parameters of the embedding layer are a weight matrix $\mathbf{W}_{emb}\in \mathbb{R} ^{d\times\mathcal{|C|}}$ and a bias vector $\mathbf{b}_{emb}\in\mathbb{R}^{d}$, where $d$ is a dimension of the continuous space. The input vector at each visit $\mathbf{x}_{t}$ is mapped. 

The 1-dimensional convolution network employs multiple filter matrices with one of their dimension fixed as the same as hidden dimension $d$, which can be denoted as $\mathbf{W}_{conv}\in\mathbb{R}^{l\times d}$. The other filter dimension $l$ denotes the size of a filter. A max pooling layer is added after the convolution operation to get the most significant dimensions formed into a vector representation for each patient. Finally, three MLP layers are used to produce the risk probabilities as a prediction $\hat{y}_i$ for the patient $i$. To sum, all of the weight matrices, as well as bias vectors in our three basic modules, make up the whole collection of parameter $\Theta$, which is optimized through feeding the network patients' data $\mathcal{D}=\{(\mathbf{X}_{i}, \mathbf{y}_{i})\}$.      

\noindent{\bf {LSTM-based Sequence Learning}}. Recurrent Neural Networks are frequently adopted as a predictive model with great promise in many sequence learning tasks. As for EHRs, RNN can in principle map from the sequential medical records of previous inputs that ``memory'' information that has been processed by the model previously. A standard LSTM model \cite{hochreiter1997long} is used to replace the convolutional layer in the CNN architecture we just introduced. LSTM weights, which are also parts of $\Theta$, can be summarized into two mapping matrix as $\mathbf{W}_{h}\in\mathbb{R}^{d\times 4d}$ and $\mathbf{W}_{x}\in\mathbb{R}^{d\times 4d}$. They are in charge of gates (input, forget, output) controlling as well as cell state updating. We keep the same network structures of the embedding layer and MLPs to make CNN and LSTM comparable for each other. 

\noindent{\bf {Learner}}. With the learned parameter $\Theta$, the prediction probability of an input matrix $\mathbf{X}_{i}$ is computed by $\hat{\mathbf{y}}_{i}=f(\mathbf{X}_{i};\Theta)$. The neural networks can be optimized by minimizing the following objective with a cross-entropy:   
\begin{equation}
\begin{aligned}
\mathcal{L}(\Theta) = - \frac{1}{N} \sum_{i=1}^{N} \left(\left(\mathbf{y}_{i}\right)^T\log\left(\hat{\mathbf{y}}_{i}\right)+\left(1-\mathbf{y}_{i}\right)^T\log\left(1-\hat{\mathbf{y}}_{i}\right)\right)
\end{aligned}
\label{eq:learner}
\end{equation}
where $N$ denotes the patient number in the training data. Similarly, the loss functions for source and target domains have the same formulation with Eq.~(\ref{eq:learner}), which are denoted as $\mathcal{L}_\mathcal{S}$ and $\mathcal{L}_{\mathcal{T}^s}$. 

\subsection{\methodname~Architecture}

\noindent{\bf {Optimization-Level Adaptation}}. In general, meta-learning aims to optimize the objective over a variety of learning tasks $\mathcal{T}$ which are associated with the corresponding datasets $\mathcal{D}_{\mathcal{T}}$. The training episodes $\mathcal{D}_{epi}$ are generated by a data distribution $p(\mathcal{D}_{\mathcal{T}})$. Then the learning procedure of parameter $\Theta$ is defined as: 
\begin{equation}
\begin{aligned}
\Theta^* = \arg \min_{\Theta} \mathbb{E}_{m}\mathbb{E}_{\mathcal{D}_{epi}^{m}\sim p(\mathcal{D}_{\mathcal{T}})}\mathcal{L}_{\Theta}(\mathcal{D}_{\mathcal{T}})
\end{aligned}
\label{eq:generalmeta}
\end{equation}
where $m$ episodes of training samples are used in the optimization. $\mathcal{L}_{\Theta}$ is the loss function that might take different formulations depending on the different strategies to design a meta-learner. As it is claimed in meta-learning, the models should be capable of tackling the unseen tasks during testing stages. In order to achieve this goal, the loss function for one episode can be further defined as the following form: 
\begin{equation}
\begin{aligned}
\mathcal{L}_{\Theta} = \frac{1}{|\mathcal{D}_{epi}^{te}|}\sum_{(\mathbf{X}_{i}, \mathbf{y}_{i})\in\mathcal{D}_{epi}^{te}}\mathcal{L}_{\Theta}\left((\mathbf{X}_{i}, \mathbf{y}_{i});\mathcal{D}_{epi}^{tr}\right)
\end{aligned}
\label{eq:epiloss}
\end{equation}

where $\mathcal{D}_{epi}^{tr}$ and $\mathcal{D}_{epi}^{te}$ are the two parts of a sample set that simulated training and testing in each episode as we introduced previously. It is worth to note that Eq.~(\ref{eq:epiloss}) is a loss decided by the prediction qualities of samples in $\mathcal{D}_{epi}^{te}$. The model-agnostic meta-learning (MAML)~\cite{finn2017model} provides us a parameter initialization scheme for $\Theta$ in Eq.~(\ref{eq:epiloss}) by taking full advantage of $\mathcal{D}_{epi}^{tr}$. It assumes that there should be some internal representations are more transferable than others, which could be discovered by an inner learning procedure using $\mathcal{D}_{epi}^{tr}$. Based on the essential idea, we show the underlying mechanism of model-agnostic meta-learning fits the problem of transferring knowledge from source domains to a low-resource target domain very well, which can be used in solving the risk prediction problem of several underdiagnosed diseases.    

Figure \ref{fig:Meta-SeqLearn} illustrates the architecture of the proposed \methodname. The general meta-learning algorithms generate episodes over task distributions and shuffle the tasks to make each task could be a part of $\mathcal{D}_{epi}^{tr}$ or $\mathcal{D}_{epi}^{te}$. Instead, we define the two disjoint parts of the episode as source domains and a target domain to satisfy a transfer learning setting. To construct a single episode $\mathcal{D}_{epi}$ in the meta-training process, we sample data via $\{(\mathbf{X}_{\mathcal{S}^i}, \mathbf{y}_{\mathcal{S}^i})\}\sim p(\mathcal{D}_{\mathcal{S}^{i}})$ and $\{(\mathbf{X}_{\mathcal{T}^s}, \mathbf{y}_{\mathcal{T}^s})\}\sim p(\mathcal{D}_{\mathcal{T}^{s}})$ respectively. In order to optimize $\Theta$ that can quickly adapt to the held-out samples in target domain, the inner learning procedure should be pushed forward by the supervise information of the source samples. To meet this requirement, the following gradient update occurs: 
\begin{equation}
\begin{aligned}
\Theta'=\Theta-\alpha\nabla_{\Theta}\sum_{i}^{K-1}\mathcal{L}_{\mathcal{S}^{i}}
\end{aligned}
\label{eq:update_source}
\end{equation}
where $\mathcal{L}_{\mathcal{S}^{i}}, i=1, \cdots, K-1$ are loss functions of source domains. $\alpha$ is a hyperparameter controlling the update rate. The source loss is computed by $\mathcal{L}_{\mathcal{S}^{i}}
=\mathcal{L}(\mathbf{y}_{\mathcal{S}^{i}}, f(\mathbf{X}_{\mathcal{S}^{i}}, \Theta)$. From Eq.~(\ref{eq:update_source}) we can observe that it is a standard form gradient descent optimization. In practice, we will repeat this process $k$ times, then output the $\Theta '$ as an initial parameter for the simulated target domain. The inner learning can be view as an Inner-Loop which is shown in Figure \ref{fig:Meta-SeqLearn}. 

Once we set $\Theta=\Theta '$ before the update step of the simulated target domain, the minimize problem defined by the loss given in Eq.~(\ref{eq:epiloss}) becomes: 
\begin{equation}
\begin{aligned}
&\min_{\Theta}\mathcal{L}_{\mathcal{T}^{s}}(f_{\Theta 
'})=
\min_{\Theta}\sum_{\mathcal{D}_{epi}^{\mathcal{T}^s}\sim p(\mathcal{D}_{\mathcal{T}^{s}})}\mathcal{L}\left(\mathbf{y}_{\mathcal{T}^{s}}, f(\mathbf{X}_{\mathcal{T}^{s}}, \Theta ')\right)
\end{aligned}
\label{eq:metaloss_target}
\end{equation}
where $\mathcal{D}_{\mathcal{T}^{s}}=\{(\mathbf{X}_{\mathcal{T}^s}, \mathbf{y}_{\mathcal{T}^s})\}$. Given the loss form of $\mathcal{L}_{\mathcal{T}^{s}}$ in the simulated target domain, it is computed by the output parameter $\Theta '$ obtain via inner gradient update in Eq. (\ref{eq:update_source}). Then, the meta-optimization using $\mathcal{D}_{\mathcal{T}^s}$ is performed with:    
\begin{equation}
\begin{aligned}
\Theta=\Theta-\beta\nabla_{\Theta}\mathcal{L}_{\mathcal{T}^s(f_{\Theta '})}
\end{aligned}
\label{eq:update_target}
\end{equation}
where $\beta$ is the meta-learning rate. Hence, the simulated target loss involves an Outer-Loop for gradient updating. Compared to the standard gradient updating in Eq.~(\ref{eq:update_source}), the gradient-like term in Eq.~(\ref{eq:update_target}) essentially resorts to a \textit{gradient through a gradient} that can be named as meta-gradient. Accordingly, the entire learning procedure can be viewed as: iteratively transfer the parameter $\Theta$ learned from source domains through utilizing it as the initialization of the parameter that needs to be updated in the simulated target domain.      

To build end-to-end risk prediction models with the model-agnostic gradient updating, we use the deep neural network structures that are trained using medical records $\mathbf{X}$ and diagnosis results $\mathbf{y}$ described in Section 3.1. The objectives for both source and simulated target are set as cross-entropy introduced in Eq.~(\ref{eq:learner}). 
One interesting point is that all the parameters of source domains and simulated target domains are tied, with different stages to update. The colors in Figure~\ref{fig:Meta-SeqLearn} provides an indication about the aforementioned two kinds of gradient pass.

\linespread{1.3}{
\begin{algorithm}[t]
\caption{\methodname~Training}
\label{alg:meta-train}
\begin{algorithmic} [1]
\renewcommand{\algorithmicrequire}{\textbf{Require:}}
\renewcommand{\algorithmicensure}{\textbf{Require:}}
\REQUIRE ~~
Source domains $\mathcal{S}^i$; Simulated target domain $\mathcal{T}^{s}$;\\
\ENSURE ~~
Hyperparameters $\alpha, \beta, \mu$;\\ 
\STATE Initialize model parameter $\Theta$ randomly 
\STATE \textbf{while} \textcolor{blue}{Outer-Loop} not done \textbf{do}\\
\STATE \quad Sample batch of episodes $\{\mathcal{D}_{epi}\}$ from $\mathcal{D}_{\mathcal{S}^{i}}$ and $\mathcal{D}_{\mathcal{T}^{s}}$\\
\STATE \quad \textbf{while} \textcolor{orange}{Inner-Loop} not done \textbf{do}\\
\STATE \quad \quad $\{(\mathbf{X}_{\mathcal{S}^i}, \mathbf{y}_{\mathcal{S}^i})\}_{i=1}^{K-1}, \{(\mathbf{X}_{\mathcal{T}^s}, \mathbf{y}_{\mathcal{T}^s})\}=\{\mathcal{D}_{epi}\}$\\
\STATE \quad \quad Compute $\mathcal{L}_{\mathcal{S}^{i}}
=\mathcal{L}(\mathbf{y}_{\mathcal{S}^{i}}, f(\mathbf{X}_{\mathcal{S}^{i}}, \Theta)), i=1,\cdots,K-1$\\

\STATE \quad \quad Parameter fast adaption with gradient descent: \\
\STATE \quad \quad $\Theta '=\Theta-\alpha\nabla_{\Theta}\sum_{i}^{K-1}\mathcal{L}_{\mathcal{S}^{i}}$\\
\STATE \quad \textbf{end for}
\STATE \quad Compute $\mathcal{L}_{\mathcal{T}^{s}}
=\mathcal{L}(\mathbf{y}_{\mathcal{T}^{s}}, f(\mathbf{X}_{\mathcal{T}^{s}}, \Theta'))$\\
\STATE \quad Update $\Theta=\Theta-\beta\nabla_{\Theta}(\mathcal{L}_{\mathcal{T}^s}+\mu\sum_{i}^{K-1}\mathcal{L}_{\mathcal{S}^{i}})$ using Adam\\
\STATE \textbf{end while} 
\end{algorithmic}
\end{algorithm}}

\noindent{\bf {Objective-Level Adaptation}}. While MAML provides an effective transferable parameter learning scheme for disease risk prediction in the low-resource situation, it cannot ensure sufficiently transferring the critical knowledge from the source domain. On the one hand, meta-learning generally encourages that the simulated target task could be randomly generated, and their model could be adapted to a large or infinite number of tasks~\cite{finn2017model, santoro2016meta, vinyals2016matching}. Different from these works, transfer learning often requires to capture domain shifts. To do so, the simulated target that is used in learning to transfer cannot be randomly sampled.   

On the other hand, the task distribution is a common decisive factor of the success for meta-learning. In other words, the distributions of the investigated source and target domains should not be too diverse. In real-world healthcare scenario, however, patients who suffering difference diseases might have medical records at various visits with heterogeneity. In this case, it is difficult to meta-learn during optimization loops. To alleviate this problem, we propose to enhance some guarantee from the objective-level in predictive modeling so that the scarcity of the fast adaptation in the optimization-level can be compensated. In particular, we propose to improve the objective by incorporating supervision from source domains. The final objective of \texttt{MetaPred} is given in the mathematical form as:               
\begin{equation}\small
\begin{aligned}
&\mathcal{L}_{\mathcal{T}}(f_{\Theta '})  = \mathcal{L}_{\mathcal{T}^{s}}(f_{\Theta 
'}) + \mu\sum_{i}^{K-1}\mathcal{L}_{\mathcal{S}^{i}}(f_{\Theta}) \\ 
&= \sum_{\mathcal{D}_{epi}^{\mathcal{T}^s}}\mathcal{L}\left(\mathbf{y}_{\mathcal{T}^{s}}, f(\mathbf{X}_{\mathcal{T}^{s}}, \Theta ')\right) 
+ \mu\sum_{i}^{K-1}\sum_{\mathcal{D}_{epi}^{\mathcal{S}^i}} \mathcal{L}\left(\mathbf{y}_{\mathcal{S}^{i}}, f(\mathbf{X}_{\mathcal{S}^{i}}, \Theta)\right)  
\end{aligned}
\label{eq:imp_metaloss_target}
\end{equation}
where $\{(\mathbf{X}_{\mathcal{S}^i}, \mathbf{y}_{\mathcal{S}^i})\}_{i=1}^{K-1}$ is a collection of medical records matrix and label vectors of source domains. $\mathcal{D}_{epi}^{T^s}$ and $\mathcal{D}_{epi}^{S^i}$ are samples from the source domain and the simulated target domain in episode $\mathcal{D}_{epi}$, respectively. Hyperparameter $\mu$ balances the contributions of the sources and simulated target in the meta-learn process. Note that the parameter of source loss is $\Theta$ but not $\Theta '$, as there is no need to conduct fast adaptation for source domain. Now the newly designed meta-gradient is updated by the following equation:       
\begin{equation}
\begin{aligned}
\Theta=\Theta-\beta\nabla_{\Theta}(\mathcal{L}_{\mathcal{T}^s}+\mu\sum_{i}^{K-1}\mathcal{L}_{\mathcal{S}^{i}})
\end{aligned}
\label{eq:imp_update_target}
\end{equation}
So far the main architecture of \methodname~is introduced. With the incorporated source loss on the basis of general meta-learning, our parameter learning process need to be redefined as:    

\begin{equation}
\begin{aligned}
\Theta^* = \mathbf{Learner}\left({\mathcal{T}^s}, \{\mathcal{S}^i\}_{i}^{K-1}; \mathbf{MetaLearner}(\{\mathcal{S}^i\}_{i}^{K-1})\right)
\end{aligned}
\label{eq:metalearn}
\end{equation}

The Algorithm~\ref{alg:meta-train} and Algorithm~\ref{alg:meta-test} are outlines of meta-training and meta-testing of the \texttt{MetaPred} framework. Similar to meta-training, episodes of the test set are consist of samples from the source domain and genuine target domain. The procedure in meta-test shows how to get a risk prediction for the given low-resource disease by a few gradient steps. The test set of the target disease domain is used to construct the meta-test episodes for the model evaluation. Since \methodname~is model-agnostic, the gradient updating scheme can be easily extended to more sophisticated neural networks including various attention mechanisms or gates with prior medical knowledge~\cite{choi2017gram, baytas2017patient}.

\linespread{1.3}{
\begin{algorithm}[t]
\caption{\methodname~Testing}
\label{alg:meta-test}
\begin{algorithmic} [1]
\renewcommand{\algorithmicrequire}{\textbf{Require:}}
\renewcommand{\algorithmicensure}{\textbf{Require:}}
\REQUIRE ~~
Source domains $\mathcal{S}^i$; target domain $\mathcal{T}^{0}$;\\
\ENSURE ~~
Learned parameter $\Theta$;\\ 
\STATE Sample from $\mathcal{D}_{\mathcal{S}^{i}}$ to construct testing episodes $\{\mathcal{D}_{epi}\}$ \\
\STATE $\{(\mathbf{X}_{\mathcal{S}^i}, \mathbf{y}_{\mathcal{S}^i})\}_{i=1}^{K-1}, \{(\mathbf{X}_{\mathcal{T}^0}, \mathbf{y}_{\mathcal{T}^0})\}=\{\mathcal{D}_{epi}\}$\\
\STATE Compute $\mathcal{L}_{\mathcal{S}^{i}} =\mathcal{L}(\mathbf{y}_{\mathcal{S}^{i}}, f(\mathbf{X}_{\mathcal{S}^{i}}, \Theta)), i=1,\cdots,K-1$\\
\STATE Parameter fast adaption with gradient descent: \\
\STATE $\Theta '=\Theta-\alpha\nabla_{\Theta}\sum_{i}^{K-1}\mathcal{L}_{\mathcal{S}^{i}}$\\
\STATE Evaluate predicted results of \textbf{Learner}$(\{(\mathbf{X}_{\mathcal{T}^0}, \mathbf{y}_{\mathcal{T}^0})\};\Theta ')$
\end{algorithmic}
\end{algorithm}}

%% file: 4-experiment.tex
\section{Experiments}

\begin{table}[t]
  \caption{Statistics of datasets with disease domains.}
  \vspace{-0.2cm}
  \label{tab:dataset}
  \begin{tabular}{|l|cc|cc|}
   \toprule
    Domain &Case &Control &\# of visit &Ave. \# of visit \\
    \midrule
    MCI &1,965 &4,388 &161,773 & 22.24 \\
    Alzheimer's &1,165 &4,628  &136,197 & 20.73  \\
    Parkinson's &1,348 &3,588 &105,053 & 20.01  \\
    \midrule
    Dementia &3,438   &1,591 &98,187 & 18.06  \\
    Amnesia &2,974     &4,215 & 180,091 & 21.60  \\
    \bottomrule
\end{tabular}
\end{table}
						
\begin{table*}[t]
\centering
\tabcolsep 0.12in
\renewcommand\arraystretch{1.25}
\caption{Performance on the disease classification tasks. The simulated target domain for three mainly investigated diseases are set as $\textbf{Alzheimer}\sim\textbf{MCI}$, $\textbf{MCI} \sim\textbf{Alzheimer}$, and $\textbf{MCI} \sim\textbf{Parkinson}$ ($\textbf{A}$ is a simulated target and $\textbf{B}$ is a target if $\textbf{A}\sim\textbf{B}$).}
\label{tab:mainresults}
\begin{tabular}{llcccccc}
\toprule
\multirow{2}{*}{Training Data}    & \multicolumn{1}{l}{\multirow{2}{*}{Model}} & \multicolumn{2}{c}{MCI} & \multicolumn{2}{c}{Alzheimer's Disease} & \multicolumn{2}{c}{Parkinson's Disease}\\ \cmidrule(lr){3-4}  \cmidrule(lr){5-6}  \cmidrule(lr){7-8}
                           & \multicolumn{1}{c}{}   &AUCROC          &F1 Score         &AUCROC        &F1 Score         &AUCROC        &F1 Score\\
\midrule
\multirow{6}{*}{Fully Supervised}  
                          &LR   &0.5861 (.01)  &0.3813 (.02)   &0.5369 (.01)   &0.2216 (.02)   &0.7504 (.01)   &0.6391 (.02) \\
                          &$k$NN  &0.6106 (.01)  &0.4540 (.01)   &0.6713 (.02)   &0.4686 (.03)   &0.7599 (.01)   &0.6403 (.01) \\
                          &RF   &0.6564 (.01)  &0.4998 (.01)   &0.6300 (.02)   &0.4111 (.04)   &0.7750 (.01)   &0.6898 (.02) \\
                          &MLP  &0.6515 (.01)  &0.5077 (.01)   &0.6639 (.02)   &0.4901 (.03)   &0.7958 (.02)   &0.7027 (.01) \\
                          &CNN  &0.6999 (.01)  &0.5816 (.02)   &0.6755 (.03)   &0.4935 (.04)   &0.7980 (.01)   &0.7265 (.02) \\
                          &LSTM &0.6874 (.01)  &0.5666 (.02)   &0.6902 (.01)   &0.5316 (.02)   &0.8041 (.02)   &0.7241 (.02) \\
\midrule
\multirow{2}{*}{Low-Resource}

                &Meta-CNN      &0.7624 (.02)   &0.6992 (.02)   &0.7682 (.01)   &0.6434 (.03)   &0.7604 (.02)   &0.6737 (.03)       \\
                &Meta-LSTM     &0.7876 (.02)   &0.7225 (.02)  &0.7464 (.02)   &0.6170 (.03)  & 0.7532 (.02)   &0.6753 (.03)     \\
\midrule
\multirow{2}{*}{Fully Fine-Tuned}
               &Meta-CNN      &0.8470 (.01)   &0.7888 (.02)   &\textbf{0.8461 (.01)}   &\textbf{0.7375 (.01)}   &\textbf{0.8343 (.01)}   &\textbf{0.7406 (.01)}       \\
               &Meta-LSTM     &\textbf{0.8477 (.01)}   &\textbf{0.7963 (.02)}  &0.8232 (.01)   &0.7364 (.01)  &0.8172 (.01)   &0.7291 (.02)     \\
\bottomrule
\end{tabular}
\end{table*}

\subsection{Dataset}
In this section, experimental results on a real-world EHR dataset are reported. The data warehouse we used in experiments is the research data warehouse (RDW) from Oregon Health \& Science University (OHSU) Hospital. The data warehouse which contains the EHR of over 2.5 million patients with more than 20 million patient encounters, is mined by Oregon Clinical and Translational Research Center (OCTRI). For certain conditions, we may not have sufficient patients for training and testing. In our study, we selected the conditions including more than $1,000$ cases (MCI, Alzheimer's disease, Parkinson's disease, Dementia, and Amnesia) as the different tasks in the multi-domain setting. For each domain, controls are patients suffering other cognitive disorders, which makes the classification tasks difficult and meaningful in practice. Also, Dementia and Amnesia are used as source domains, while the more challenging tasks MCI, Alzheimer, Parkinson are set as target domains.


We matched the case and controls by requiring their age difference within a 5-year range so that the age distributions between the case group and control group are consistent. 
For each patient, we set a 2-year observation window to collect the training data, and the prediction window is set to half a year (i.e., we are predicting after half a year the onset risk of those conditions). In our experiments, only patient diagnoses histories are used, which include 10,989 distinct ICD-9 codes in total. We further mapped them to their first three digits, which ends up with 1,016 ICD-9 group codes. The data statistics are summarized in Table~\ref{tab:dataset}.

\subsection{Experimental Setup}

\noindent{\bf {Metric \& Models for comparison}}. In our experiments, we take the \textsl{AUROC} (area under receiver operating characteristic curve) and \textsl{F1 Score}
as the prediction performance measures. 
We compare the performance of the \texttt{MetaPred} framework with the following approaches established on the target task.


\begin{figure*}[t]
\subfigure[MCI]{
\begin{minipage}[b]{0.32\linewidth}
\centering 
\includegraphics[width=\textwidth]{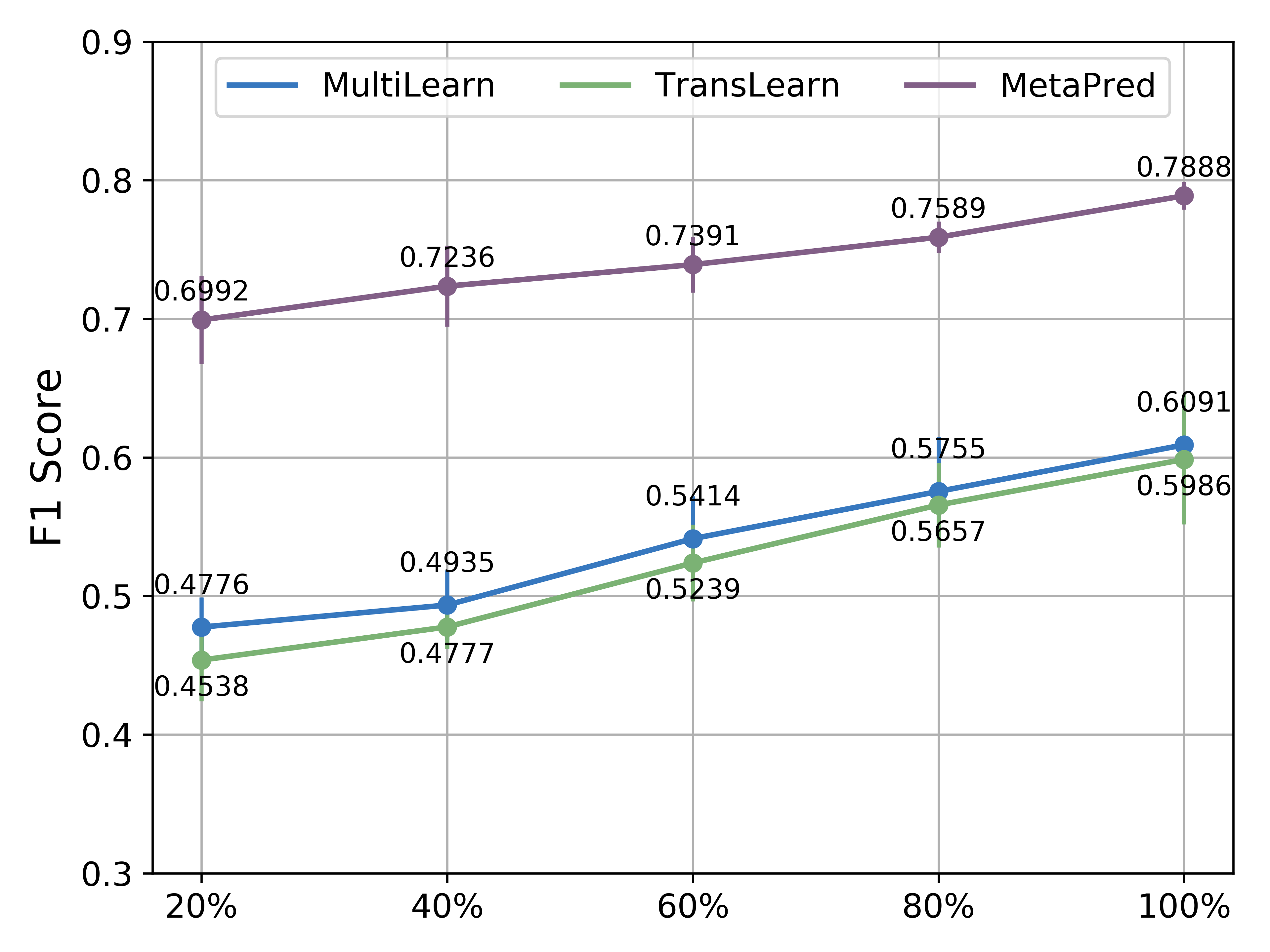}
\end{minipage}}%
\subfigure[Alzheimer's Disease]{
\begin{minipage}[b]{0.32\linewidth}
\centering
\includegraphics[width=\textwidth]{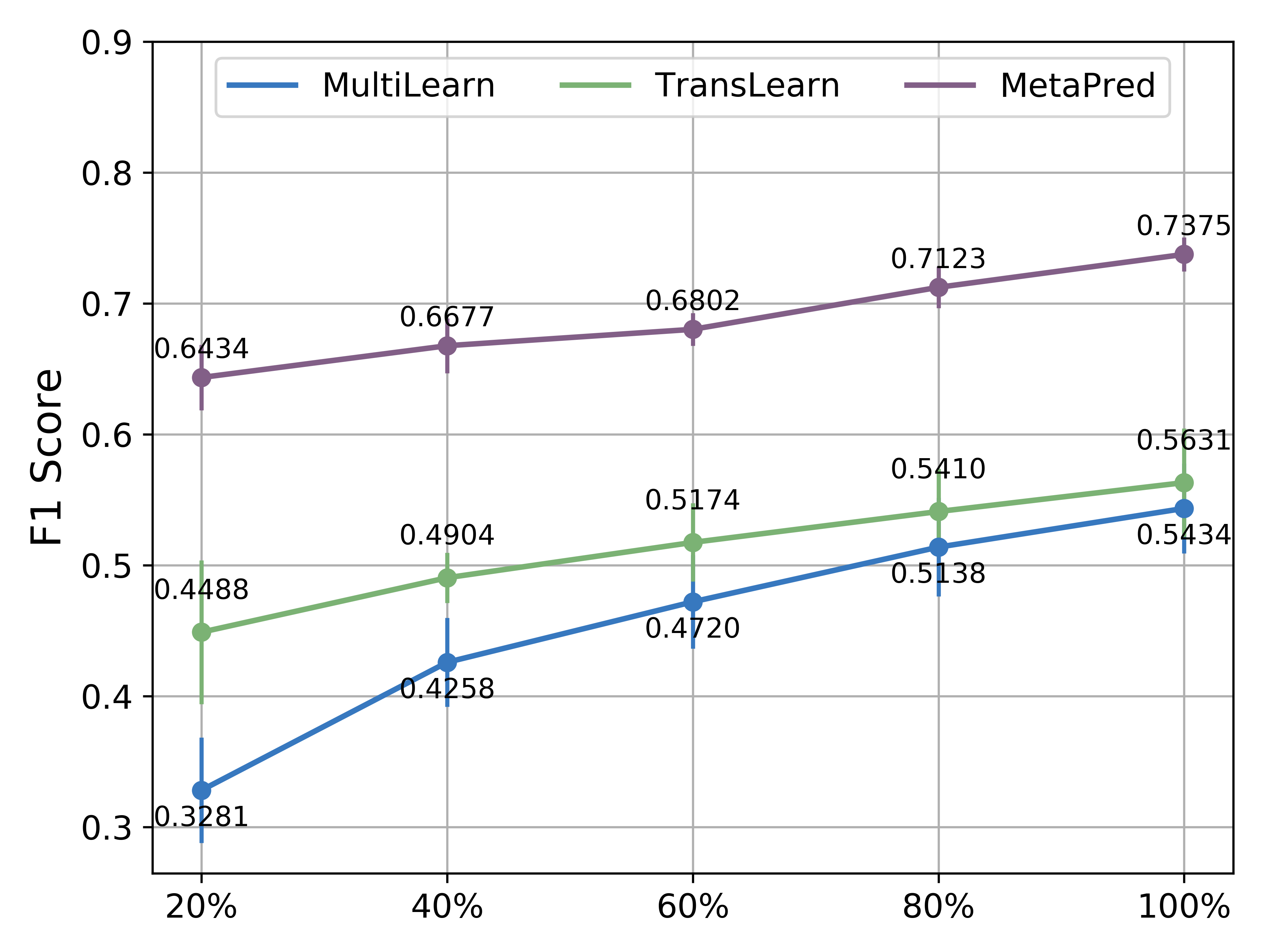}
\end{minipage}}
\subfigure[Parkinson's Disease]{
\begin{minipage}[b]{0.32\linewidth}
\centering
\includegraphics[width=\textwidth]{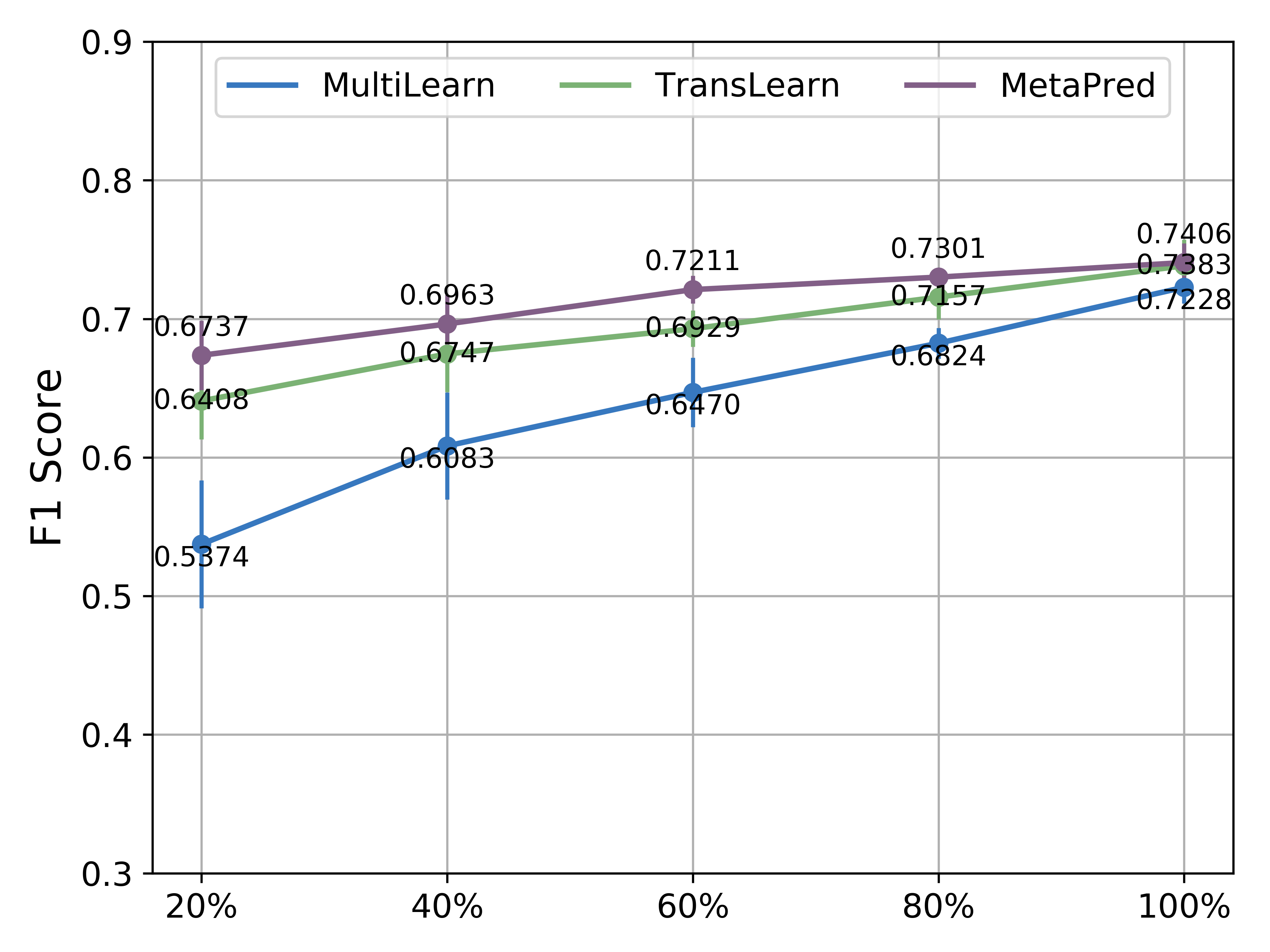}
\end{minipage}}
\vspace{-0.5cm}
\caption{Results with respect to different levels of labeled data resource used in fine-tuning for target domains.}
\label{fig:fine-tune}
\end{figure*}

\noindent\underline{\emph{Supervised classification models.}} 
Three traditional classification models without considering any sequential EHR information, including Logistic Regression (LR), $k$-Nearest Neighbors algorithm ($k$-NN), and Random Forest (RF), are implemented as baselines, where the patient vectors are formed by counting the frequencies of specific diagnosis codes during the observation window. Deep learning models, including \texttt{Embedding Layer-MLP} and \texttt{Embedding Layer-CNN/LSTM-MLP} architectures are implemented as baselines. 


\noindent{\underline{\emph{Fine-tuned models.}}} 
For the adaptation to a target domain, training data of target domains can be used in fine-tuning an established meta-learning model based on sources.  Among the basic blocks of the built networks, we consider fine-tuning MLP layers meanwhile freeze the embedding layer and CNN/LSTM blocks. Therefore, MLP can be viewed as a task-specific architecture leaned based on the corresponding target.  

\noindent{\underline{\emph{Low-Resources models.}}}
Since there are no prior efforts focusing on the critical problem of low-resource medical records. We propose two variants of \texttt{MetaPred} to verify its feasibility and effectiveness. Depends on the choice of modules for sequencing learning, we build Meta-CNN and Meta-LSTM to predict disease risks with limited target samples. Specifically, patients in the true target domain are not used in generating the episodes during meta-training, which makes our setup satisfying the meta-learning tasks. Then a small part of the training target set is employed to fine-tune the learned models. We keep this ratio as $20\%$ to simulate low-resource situations. 

To show the superior of the parameter transferable ability, we compare the performance of \texttt{MetaPred} with a basic parameter transfer learning algorithm~\cite{pan2010survey, lawrence2004learning}, which solves the following posterior maximization problem:
\begin{equation}
\begin{aligned}
\arg \max_{\Theta} \Sigma_{(\mathbf{X}, y) \in \mathcal{D}_{\mathcal{T}}} \log p(y|\mathbf{X}, \Theta)-\gamma \|\Theta-\Theta^{0}\|
\end{aligned}
\label{eq:trans}
\end{equation}
where $\Theta^{0}$ is an initial parameter setting for the target domain. The norm term gives a prior distribution of parameters and constraints that the learned model for target task should not deviate too much from the one learned from source tasks. 
The transfer learning models are named \texttt{TransLearn}. In addition, multitask learning methods~\cite{collobert2008unified, caruana1997multitask} are employed to be another comparison in the limited-resource scenario. In particular, we fix the bottom layers and use domain-specific MLP in the multitask baseline \texttt{MultiLearn}. For a fair comparison, the above approaches are all evaluated by held-out test sets of the target domains.      

\noindent{\bf {Implementation Details and Model Selection}}. For all above algorithms, 20\% patients of the labeled patients are used as a test set for the three main tasks and train models on the remaining 80\%. We randomly split patients with this ratio for target domain and run experiments five times. The average performance is reported. The deep learning approaches including the proposed \texttt{MetaPred} are implemented with Tensorflow. The network architectures of CNN and LSTM, as well as other hyperparameters are tuned by the $5$-fold cross-validation. In detail, The hidden dimensions of embedding layer and $2$ fully connected layers are set as $d_{emb}=256$ and $d_{mlp}=128$. The vocabulary size is consistent with ICD-9 diagnosis codes, which is grouped as $d_{vol}=1017$ including $1$ padding index. The sequence length is chosen according to the average number of visit per patient in Table \ref{tab:dataset}. Batch normalization~\cite{ioffe2015batch} and layer normalization~\cite{ba2016layer} are employed based on CNN and LSTM respectively. We keep the same network configurations for single task models and meta-learning models. 
We use Adam \cite{kingma2014adam} optimizer with a batch size of $32$ episodes to compute the meta-gradient. In each episode, the number of patients used for each domain is set at $8$. The proposed \texttt{MetaPred} is trained on machines with NVIDIA TESLA V100 GPUs. The source code of \texttt{MetaPred} is publicly available at~\url{https://github.com/sheryl-ai/MetaPred}.  

\subsection{Performance Evaluation}

\noindent{\bf {Performance on Clinical Risk Prediction}}.
The performance of compared approaches on three mainly investigated risk prediction tasks are presented in Table~\ref{tab:mainresults}. According to how many training data used in the target domain, there are full supervised baselines including traditional classifiers and deep predictive models, our proposed methods Meta-CNN/LSTM partially using the training data in fine-tuning, as well as the fully fine-tuned \texttt{MetaPred} models. The medical knowledge about cognitive diseases suggests us that MCI and Alzheimer are fairly difficult to be distinguished with other relevant disorders. Nevertheless, the symptoms of Parkinson's Disease sometimes are obvious to be recognized, which makes it a relatively easier task. 

From Table~\ref{tab:mainresults} we can observe that results obtained by LR, $k$NN, RF, and neural networks cannot achieve a satisfying classification performance through merely modeling the target tasks of MCI and Alzheimer. Our method Meta-CNN/LSTM perform better than single task models even with only $20\%$ labeled target samples in fine-tuning. The AUC of \texttt{MetaPred} reaches at $0.7876\pm.02$ and $0.7682\pm.01$ while their corresponding single-task versions only have $0.6874\pm.01$ and $0.6755\pm.03$. As for Parkinson, because of the insufficient labeled data, the results of low-resource cannot beat CNN/LSTM. It also indicates that the domain shift exists in real-world disease predictions. Under the fully fine-tuned setting, the labels of targets are the same as the fully supervised setting. \texttt{MetaPred} achieves significant improvements on all the three classification tasks in terms of AUC and F1 Score.           

\noindent{\bf {Comparisons at the different resource levels}}.
In order to show the superiority of \texttt{MetaPred} in the transferability with multiple domains, transfer learning and multitask learning methods are used in comparisons. Figure~\ref{fig:fine-tune} shows F1 Score results giving labeled targets samples at the percentage \{$20\%, 40\%, 60\%, 80\%, 100\%$\} of the available training data in target domain. For the transfer learning model \texttt{TransLearn} in Eq.(\ref{eq:trans}), we tried various tasks as source domains and finally used the setting  $\textbf{Alzheimer} \sim\textbf{MCI}$, $\textbf{MCI} \sim\textbf{Alzheimer}$, and $\textbf{MCI}\sim\textbf{Parkinson}$ where the best performance achieved. Meanwhile, \texttt{MultiLearn} models are compared with the same level of supervision in the three given target tasks. We randomly picked the labeled data from training set five times, and the mean and variance are presented in Figure~\ref{fig:fine-tune}. We adopt CNN as the predictive model for the compared methods here. As we can see, \texttt{MetaPred} outperforms \texttt{TransLearn} and \texttt{MultiLearn} on all of the tasks. The gap is large for MCI and Alzheimer especially when the labeled data are low. The \texttt{TransLearn} method can also perform well on the Parkinson task due to their homogeneity in several symptoms. Overall, the fast adaptation in both optimization-level and objective-level leads to more robust prediction results under low-resource circumstances.

\begin{figure}[t]
\subfigure[Meta-CNN vs. MAML-CNN]{
\begin{minipage}[b]{0.5\linewidth}
\centering 
\includegraphics[width=\textwidth]{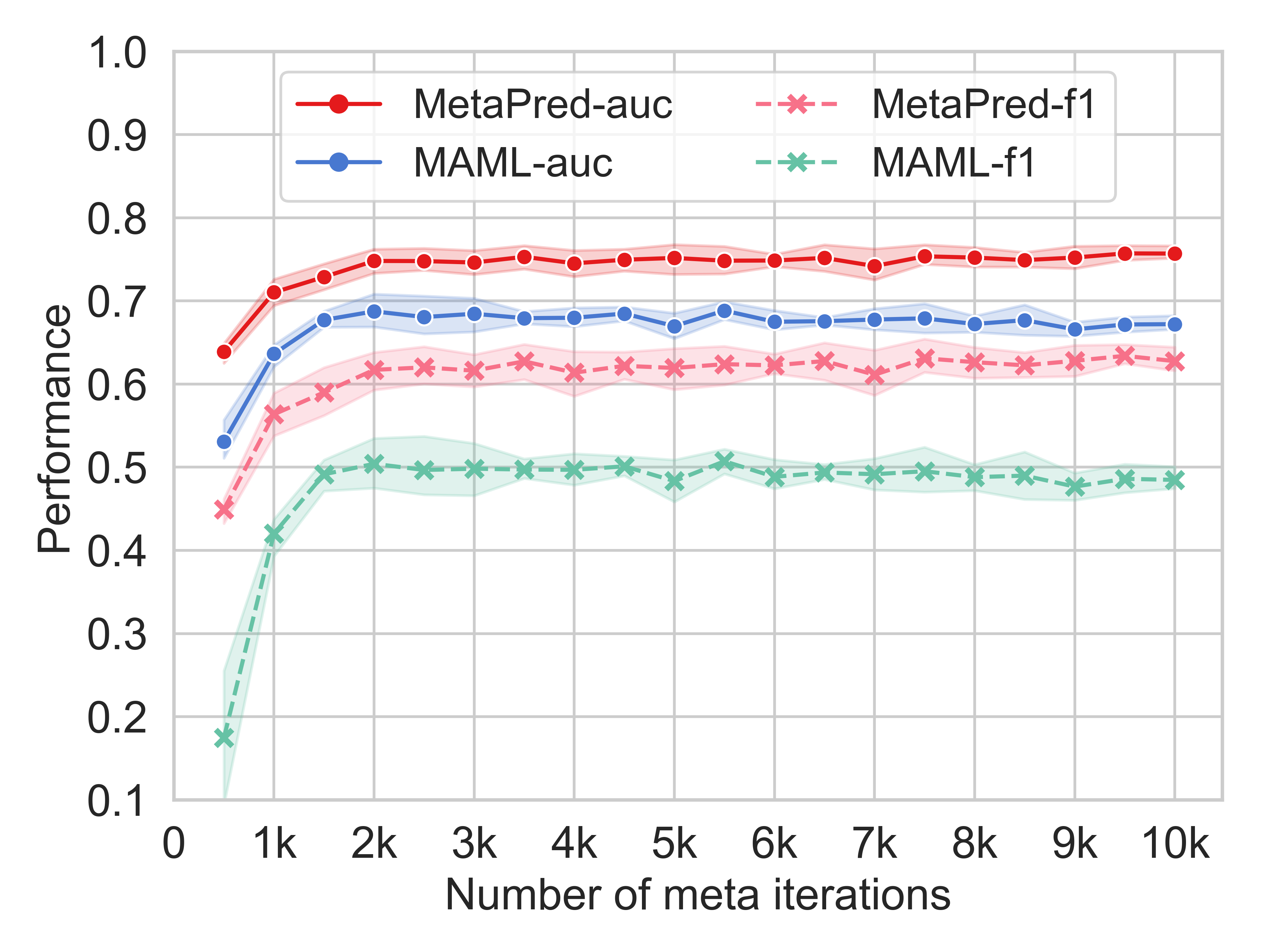}
\end{minipage}}%
\subfigure[Meta-LSTM vs. MAML-LSTM]{
\begin{minipage}[b]{0.5\linewidth}
\centering
\includegraphics[width=\textwidth]{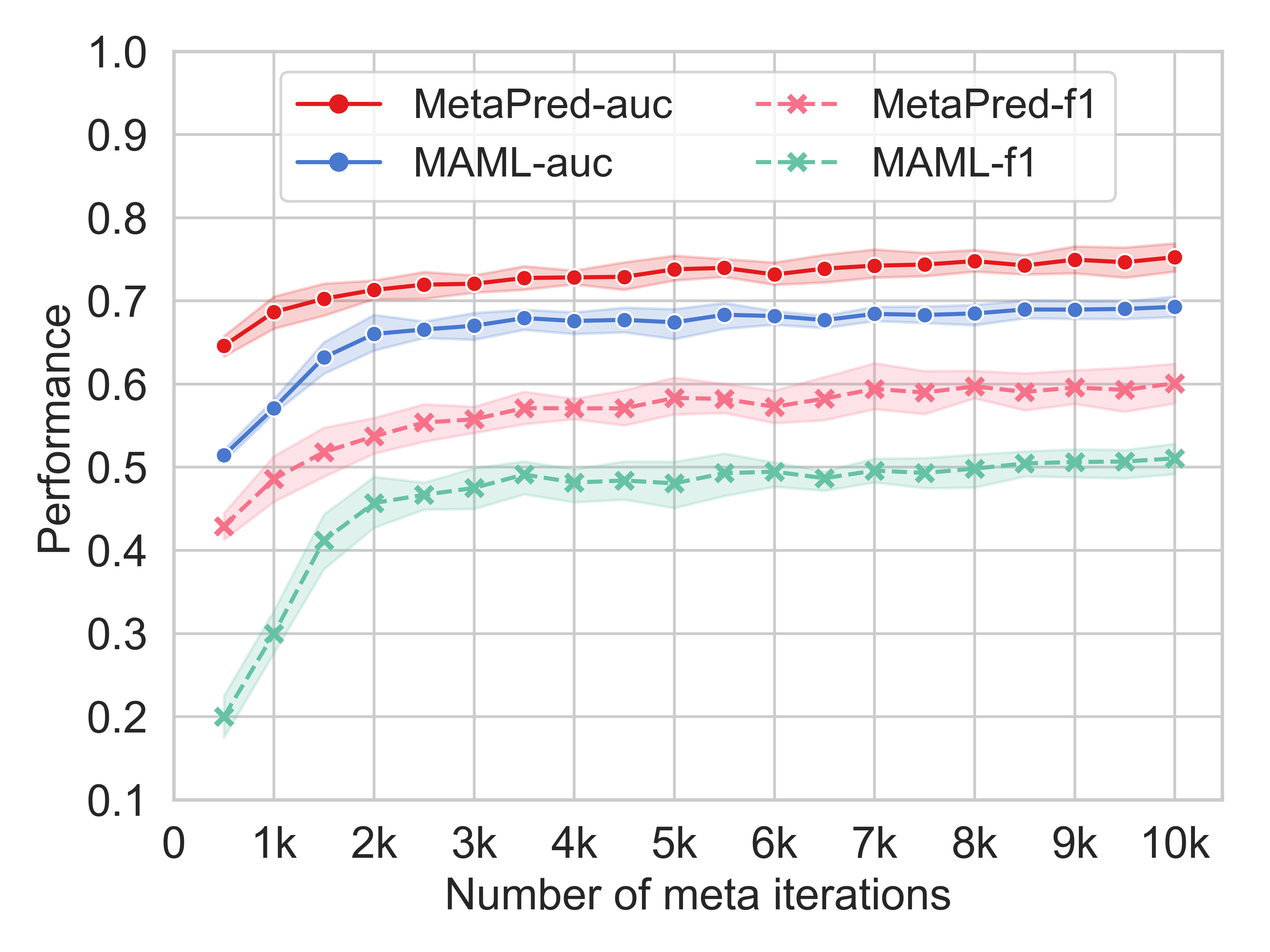}
\end{minipage}}
\caption{Comparison between \texttt{MetaPred} and MAML in terms of performance curve along with the learning procedures (Results on Alzheimer's Disease).}
\label{fig:vs_MAML}
\vspace{-0.5cm}
\end{figure}

\begin{figure*}[t]
\subfigure[MCI]{
\begin{minipage}[b]{0.32\linewidth}
\centering 
\includegraphics[width=\textwidth]{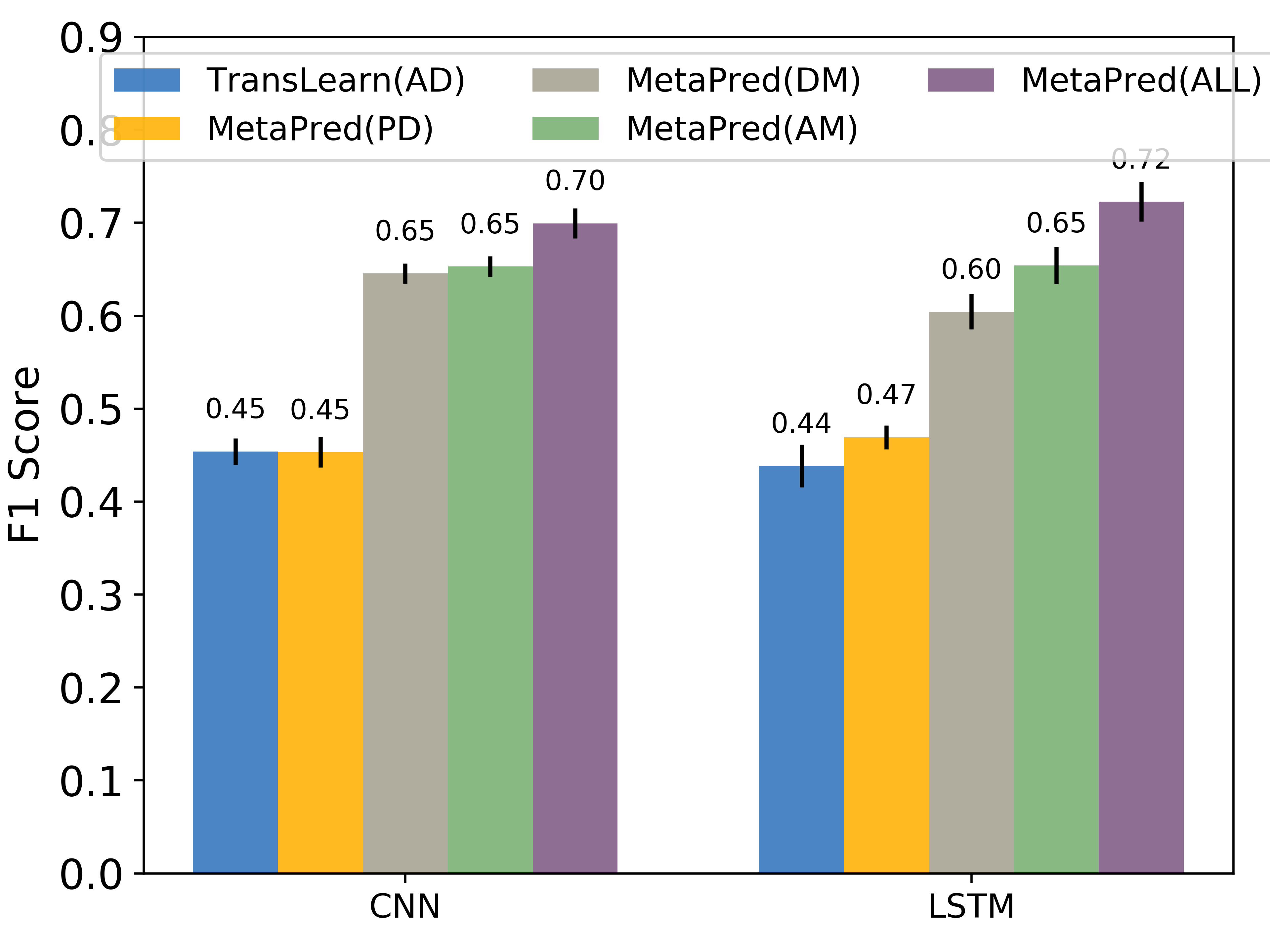}
\end{minipage}}%
\subfigure[Alzheimer's Disease]{
\begin{minipage}[b]{0.32\linewidth}
\centering
\includegraphics[width=\textwidth]{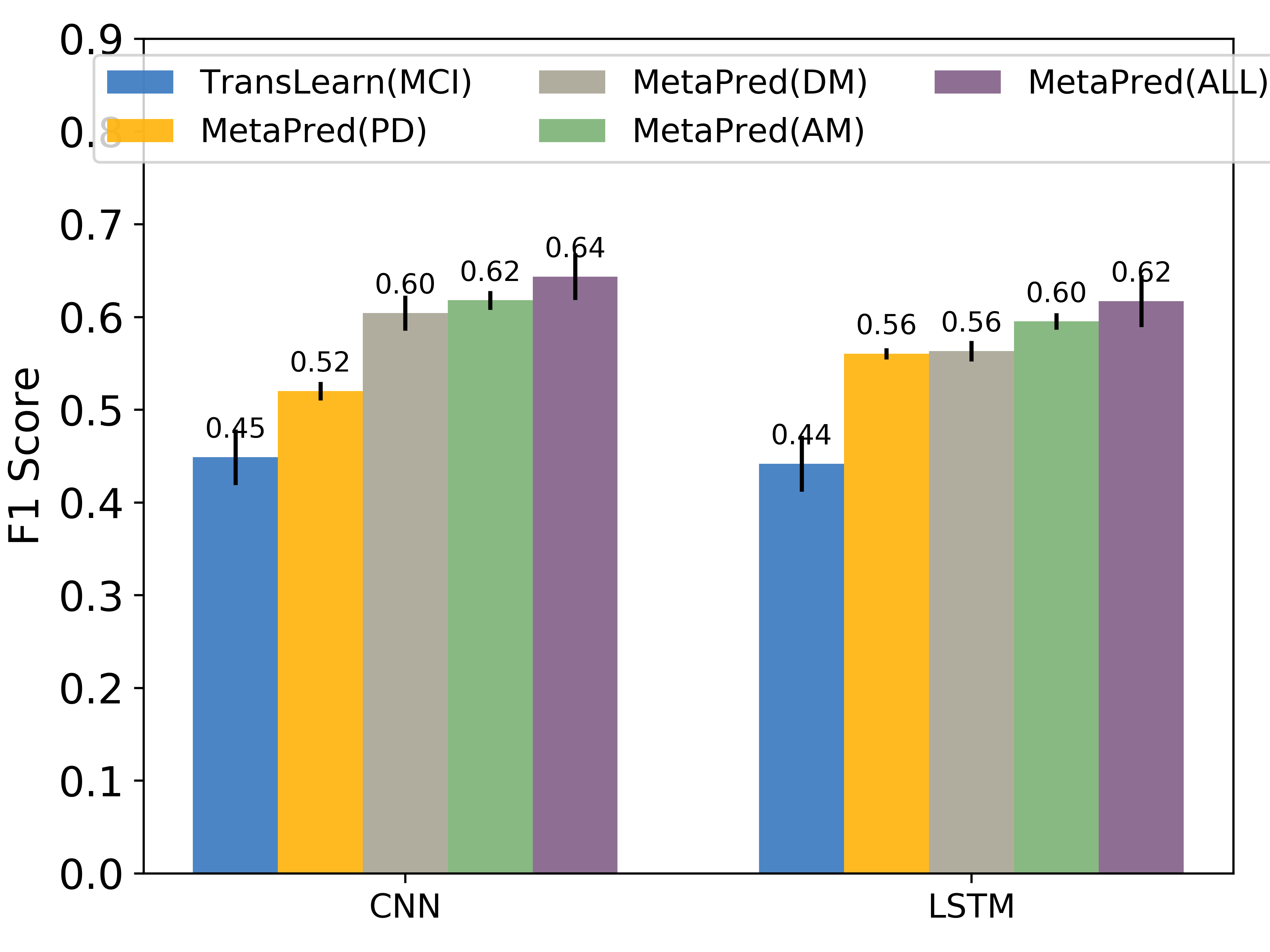}
\end{minipage}}
\subfigure[Parkinson's Disease]{
\begin{minipage}[b]{0.32\linewidth}
\centering
\includegraphics[width=\textwidth]{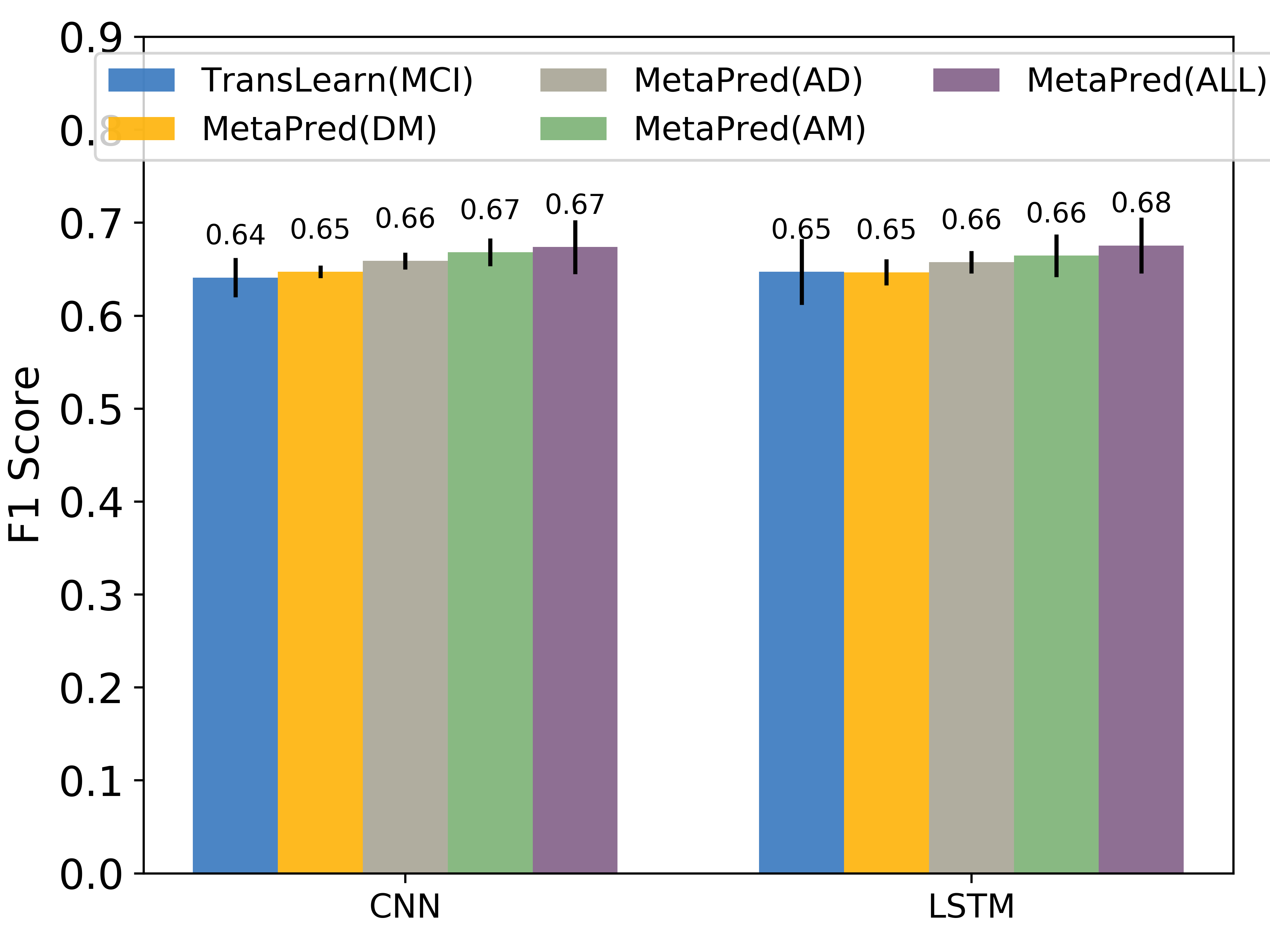}
\end{minipage}}
\vspace{-0.2cm}
\caption{Results with respect to different combinations of source disease domains. The best results among different source domains are reported for the transfer learning method (Compared methods are all under the low-resource setting).}
\label{fig:source_domain}
\end{figure*}


\begin{figure}[t]
    \centering
    \includegraphics[width=2.1in]{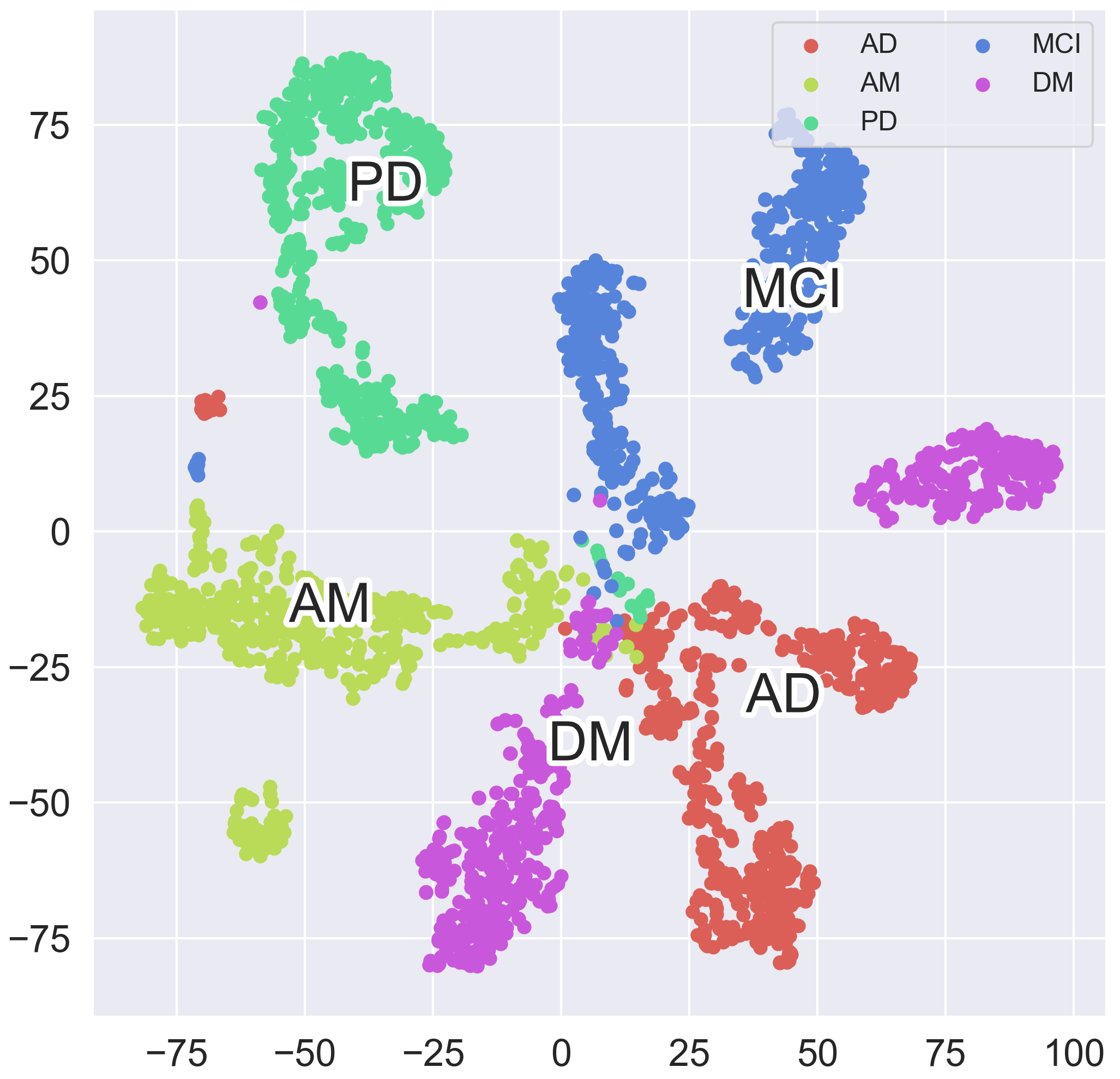}
    \caption{Visualization using a t-SNE plot of patient representation in a $2$ dimensional space. Node denotes patient suffering cognition disorders we studied. Color indicates the associated domains.}
    \label{fig:vis}
\vspace{-0.2cm}    
\end{figure}

\noindent{\bf {\texttt{MetaPred} vs. MAML}}.
To demonstrate the effectiveness of the proposed \texttt{MetaPred} learning procedure and to empirically certify the rationality of objective-level adaptation, we compare it with the state-of-the-art meta-learning algorithm MAML~\cite{finn2017model}. Experimental results of this comparison are shown in Figure~\ref{fig:vs_MAML}. To simulate the low-resource scenario, both \texttt{MetaPred} and MAML use all the available samples from sources and a simulated target for meta-train and $20\%$ labeled target patients in fine-tuning. To make the comparison fair, we use the same sets of labeled patients in the evaluation. The experiments are repeated five times, and the averaged performance with the confidence interval set as $95\%$ are given. Figure~\ref{fig:vs_MAML} gives results in terms of AUC and F1 Score for Alzheimer's Disease classification using both CNN and LSTM as the base predictive models. Along with the training iterations, the metric scores of both \texttt{MetaPred} and MAML converge to a stable value, suggesting the stability of the meta-optimization. Our method \texttt{MetaPred} achieve better performance in the disease risk prediction tasks by incorporating the supervised information of source domain.

\noindent{\bf {Impact of Source Domain}}. In Figure~\ref{fig:source_domain}, we vary the source domains as \{DM, PD, AM\}, \{DM, PD, AM\}, and \{AD, DM, AM\}\footnote{AD, PD, DM, AM are abbreviations of Alzheimer's Disease, Parkinson's Disease, Dementia, and Amnesia, respectively.} and show the F1 Score results for MCI, Alzheimer, and Parkinson, respectively. \texttt{TransLearn} is used as a baseline here. Similarly, the simulated targets are set as $\textbf{Alzheimer} \sim\textbf{MCI}$, $\textbf{MCI} \sim\textbf{Alzheimer}$, and $\textbf{MCI} \sim\textbf{Parkinson}$. Once the simulated target is fixed, we first evaluate the source domain one-by-one, then feed all of them through episode generator in meta-train. Compared to \texttt{TransLearn}, the variants of \texttt{MetaPred} generally performs better on the basis of both CNN and LSTM. Intuitively, using samples from more source domains leads to a more comprehensive representation space and thus a better prediction result on targets, which is verified by Figure~\ref{fig:source_domain} very well. Besides, source domains have an influence on the performance largely, especially for MCI and Alzheimer. For example, the largest gap of F1 Score could be close to $0.25$ in MCI prediction. The analysis helps us to choose the source domain according to their performance on the target predictions. That is, Amnesia always benefits more as a source domain whereas Parkinson benefits less compared to other sources.

\noindent{\bf {Visualization}}. Figure~\ref{fig:vis} provides the visualization results. The representations learned before the last MLP layer of \texttt{MetaPred} can be extracted as high-level features for patients. The feature dimension is $128$ as we aforementioned. During the representation learning, we hold-out $518$ cases from each domain, and build a \texttt{MetaPred} upon the rest of the data. Then, the held-out patients are clustered via t-SNE based on the outputted representations. It is shown that the five diseases are separated quite well and suggests that \texttt{MetaPred} generates meaningful representations for patients in several relevant domains.                


%% file: 5-relatedwork.tex
\section{Related Work}

Meta-learning, also known as learning to learn \cite{andrychowicz2016learning, thrun1998learning,  lake2015human}, aims to solve a learning problem in the target task by leveraging the learning experience from a set of related tasks. 
Meta-learning algorithms deal with the problem of efficient learning so that they can learn new concepts or skills fast with just a few seen examples. Meta-learning algorithms have been recently explored on a series of topics including few-shot learning \cite{ravi2016optimization, vinyals2016matching, santoro2016meta}, reinforcement learning \cite{finn2017model, ritter2018been} and imitation learning \cite{finn2017one}. One scheme of meta-learning is to incorporate learning structures of data points by distance functions \cite{koch2015siamese} or embedding networks \cite{vinyals2016matching, snell2017prototypical} such that the classifier can adapt to accommodate unseen tasks in training. Another scheme is basically optimization-based which is training a gradient procedure and applied it on a learner directly \cite{ravi2016optimization, andrychowicz2016learning, finn2017model, santoro2016meta}. Both of the schemes could be summarized as the design and optimization of a function $f$ which gives predictions for the unseen testing data $\mathbf{X}_{test}$ with training episodes $\mathcal{D}_{epi}$ and parameter collection $\Theta$. Specifically, model-agnostic meta-learning \cite{finn2017model} aims to learn a good parameter initialization for the fast adaptation of testing tasks. It has gained successes in applications such as robotic \cite{clavera2018learning, finn2017one} and neural machine translation \cite{gu2018meta}.

However, the application of meta-learning in healthcare has rarely been explored, despite the fact that most of the medical problems are resource-limited. Consequently, we propose \texttt{MetaPred} to address the general problem of clinical risk predictions with low-resource EHRs.   


%% file: 6-conclusion.tex
\section{Conclusion} 

In this paper, we propose an effective framework \texttt{MetaPred} that can solve the low-resource medical records problem in clinical risk prediction. \texttt{MetaPred} leverages deep predictive modeling with the model agnostic meta-learning to exploit the labeled medical records from high-resource domain. For the purpose of designing a more transferable learning procedure, we introduce a objective-level adaptation for \texttt{MetaPred} which not only take advantage of fast adaptation from optimization-level but also take the supervision of the high-resources domain into account. Extensive evaluation involving $5$ cognitive diseases is conducted on real-world EHR data for risk prediction tasks under various source/target combinations. Our results demonstrated the superior performance of \texttt{MetaPred} with limited patient EHRs, which can even beat fully supervised deep neural networks for the challenging risk prediction tasks of  MCI and Alzheimer.  For future clinical study, comprehensive longitudinal records more than $5$ years will be explored for cognition related disorders.

%% file: 7-acknowledgment.tex
\section*{Acknowledgement}

The research is supported by NSF IIS-1750326, ONR N00014-18-1-2585, Oregon Alzheimer's Disease Center and Michigan AD center grants NIH P30AG008017 and NIH P30AG053760.